\documentclass[fonts]{icst}

\usepackage{moreverb}

\usepackage[breaklinks,colorlinks,bookmarksopen,bookmarksnumbered,linkcolor=ICSTblue,citecolor=blue,urlcolor=ICSTblue]{hyperref}
\usepackage{breakurl}
\usepackage{doi}
\usepackage{amsmath}
\usepackage{verbatim}
\usepackage{graphicx}
\usepackage{url}
\usepackage{listings}
\usepackage{algorithmic}
\usepackage{natbib}
\usepackage{pslatex} % Use Postscript fonts
\usepackage{multirow}
\usepackage{array}

\newcommand\BibTeX{{\rmfamily B\kern-.05em \textsc{i\kern-.025em b}\kern-.08em
T\kern-.1667em\lower.7ex\hbox{E}\kern-.125emX}}

\def\volumeyear{2016}

\begin{document}

\runningheads{H. Ar Rosyid, M. Palmerlee, K. Chen}{A case study of deploying learning materials into game content}

\title{Deploying learning materials to game content for serious education game development: A case study}

\author{Harits Ar Rosyid\affil{1}\fnoteref{1}, Matt Palmerlee\affil{2}, Ke Chen\affil{1}}
%\author{Harits Ar Rosyid\affil{1}, Matt Palmerlee\affil{2}, Ke Chen\affil{3}}

\address{\affilnum{1}School of Computer Science, The University of Manchester\\
\affilnum{2}Engineering at Mastered Software, Los Angeles}

\abstract{The ultimate goals of serious education games (SEG) are to facilitate learning and maximizing enjoyment during playing SEGs.  In SEG development, there are normally two spaces to be taken into account: knowledge space regarding learning materials and content space regarding games to be used to convey learning materials. How to deploy the learning materials seamlessly and effectively into game content becomes one of the most challenging problems in SEG development. Unlike previous work where experts in education have to be used heavily, we proposed a novel approach that works toward minimizing the efforts of education experts in mapping learning materials to content space. For a proof-of-concept, we apply the proposed approach in developing an SEG game, named \emph{Chem Dungeon}, as a case study in order to demonstrate the effectiveness of our proposed approach. This SEG game has been tested with a number of users, and the user survey suggests our method works reasonably well.}


%\tnotetext[1]{Please ensure that you use the most up to date
%class file, available from EAI at \url{http://doc.eai.eu/publications/transactions/latex/}
%}

\fnotetext[1]{Corresponding author.  Email: \email{harits.arrosyid@manchester.ac.uk}}

\maketitle
\section{Introduction}

Serious Educational Game (SEG) refers to an alternative learning methodology that applies game technology to primarily promoting players' learning along with gaining positive cognitive and affective experience during such a learning process \cite{m_chen:sg}. Elements of challenge and learning within such a game construct activities for motivation and amusement~\cite{gee2005good}. SEG is also named in different terminologies such as game-based learning or educational games. In this paper, we treat all those terminologies interchangeably and refers the SEG development to the procedure that builds up a game for a learning purpose.

\par
There are useful approaches to game development for a learning purpose, such as \cite{freitas2006framework,ibrahim2009educational}. Most of those approaches emphasize that the design of a serious game is mainly from learning materials in a subject. Moreover, the proposed development frameworks require rigorous procedures that may involve interviews with target users (including teachers and students) and various experts (e.g. game development, education, psychology and so on), lengthy development stages and testing units. Hence, those development frameworks have to rely on a close relationship between learning materials and game design (proprietary educational game). Such development frameworks inevitably incur the high cost as this development process is laborious and time-consuming and hence limit the growth of educational games.

\par
In general, SEG development has to involve two key components: \emph{knowledge} and \emph{game content} spaces \cite{ismailovic2012adaptive,hussaan:tailor}.  The knowledge space is formed to encode learning materials concerning the subject knowledge to be learned by players, while the game content space is formed with playable game elements that convey the knowledge chunks implicitly. This is generally required by any serious games as argued in \cite{SG:djaoutiClassifying, SG:alvarez2011introduction} where serious game is defined as a computer program that combines \textit{serious} (for knowledge learning ) and \textit{game} (for entertainment) purposes. Thus, how to map the knowledge space to content space becomes one of the most important problems in SEG development.
To our knowledge, however, the mapping is a bottle-neck in SEG development as this has to be handcrafted by game developers closely working with education experts in most of existing SEGs.

Unlike most of existing approaches, we propose an alternative SEG development framework in this paper to address the mapping issue by embedding annotated knowledge chunks into categorized game content/elements seamlessly during SEG development. On one hand, there are abundant educational resources (e.g. syllabus) that provides the connection between underlying knowledge chunks as well as sufficient instruction~\cite{ibrahim2009educational} for learning them. Our framework would exploit such information so that knowledge chunks and their connections can be easily annotated by game developers or automatically acquired by using information retrieval techniques. On the other hand, the ``purpose-shifting'' approach has been proposed for SEG development \cite{gee2003video, SG:shaffer2006computer}, which diverts the purpose of an existing commercial game for educational propose.  This approach exploits the education-related  properties of existing commercial games, e.g., in order to play a game, a player has to learn game rules, objectives and strategy to success unconsciously, which is also required for learning in traditional education systems. As an alternative game development methodology, procedural content generation can generate game content automatically via algorithms, which significantly lowers the cost in game development. Moreover, the latest PCG work \cite{robert2013learningbased} suggests that a proper use of the categorized game content may facilitate eliciting positive gameplay experience. Motivated by the aforementioned works, our framework would advocate making use of PCG and existing entertainment games in SEG development. In particular, we believe that the mapping between two spaces may better done by embedding annotated knowledge chunks into categorized game content/elements.

\par
The main contributions of the work presented in this paper are summarized as follows: a) we propose an alternative framework for effective and efficient SEG development; b) under our proposed framework, we develop a proof-of-concept SEG, Chem Dungeon, to demonstrate the usefulness of our proposed framework; and c) we test this SEG with human players via user survey and statistical analysis.

\par
The rest of the paper is organized as follows. Section 2 reviews the related works. Section 3 presents our SEG framework and Section 4 describes our proof-of-concept SEG, Chem Dungeon. Section 5 reports user test and statistical analysis results. The last section draws conclusions.

\section{Related Work}
In this section, we outline connections and main differences to relevant SEG development approaches.

\par
As argued by Damir et al.~\cite{ismailovic2012adaptive} based on their interview with education experts, game developers and players who involve in SEG, it is crucial to have seamless connection between knowledge and game content spaces in SEG development. Moreover, they further emphasize that two spaces must be controllable \cite{ismailovic2012adaptive} to allow for gaining the controllability in tailoring game elements that are likely affecting different kinds of the player's experience, e.g. learning, enjoyment, motivation, engagement and so on. Moreover, it is suggested by Hussaan et al. \cite{hussaan:tailor} that there are three components in SEG. Apart from learning and game resources,  domain concept should be introduced to specify the relationships between learning materials to facilitate using learning resources to formulate strategies in carrying out learning based on game resources. Nevertheless, this approach \cite{hussaan:tailor} emphasizes that all of those components have to be carried by education experts via interactions with students or game players.

\par
Gamification ~\cite{SG:kapp2012gamification} is a typical SEG development approach that explicitly takes knowledge and game content spaces into account in development. The basic idea underlying gamification is directly embedding game elements (e.g. avatar, badges, levels and scores) into the learning process. Doing so expects that students would more actively engaged in the learning process when they are situated in a game-like presentation of the learning materials. In this work ~\cite{SG:kapp2012gamification},  the connection between two spaces has been handcrafted by both education experts and game developers, which is laborious and time consuming. Similarly, Belloti et al. proposed a generic approach for adaptive experience in serious games via building up the proper connection between knowledge and game content spaces ~\cite{belloti:xeng}. In their approach, a serious game is manually broken down into a number of  subsequent tasks by considering diversified connections between learning materials and game elements. Then adaptation is carried out by offering a proper task sequence to an individual player to maximize their positive learning and positive affective experience~\cite{belloti:xeng}. However, the game design (in particular, mapping between two spaces) has to heavily rely on education experts, and it is infeasible to develop such serious games without involving education experts. Due to the use of education experts, the cost in serious game development is often very high. Technically, such an approach is also subject to limitation since the mapping task becomes extremely challenging when either of two spaces is of a high complexity. Hence, this approach is not scalable in SEG development.

\par
Unlike the above approaches, our proposed SEG framework would exploit the existing educational resources and make use of appropriate PCG techniques towards minimizing the cost and seamlessly connecting knowledge and game content spaces in SEG development.

\section{Methodology} \label{sec:method}

In this section, we propose an alternative framework for SEG development especially for addressing the mapping issue pertaining to two spaces via exploiting learning resources and making use of the latest PCG techniques.

\par
The advantage of structuring serious game content in knowledge and game content spaces provides a higher degree of control for the game generation. In the existing SEG approaches,
however, education experts have to be the prominent force in the process of deploying learning materials into an SEG. Thus, an expert is expected to deeply understand characteristics of learning materials and game content according to their expertise in order to link two spaces. However, it becomes infeasible and scalable in presence of complex game content space. Hence, game developers are expected to utilize the natural and inherent game elements to deal with the knowledge deployment issue. This is feasible since sophisticated education resources are accessible easily and the PCG techniques allow for flexibly controlling game elements to embed knowledge chunks. Thus, we believe that exploiting learning resources and making use of the latest PCG techniques could significantly lower the cost of SEG development; given the semantic descriptions of those content spaces, the developer can formulate different aspects between them, which sparks a proper deployment.

\par
\begin{figure}
\centering
\includegraphics[scale=0.5]{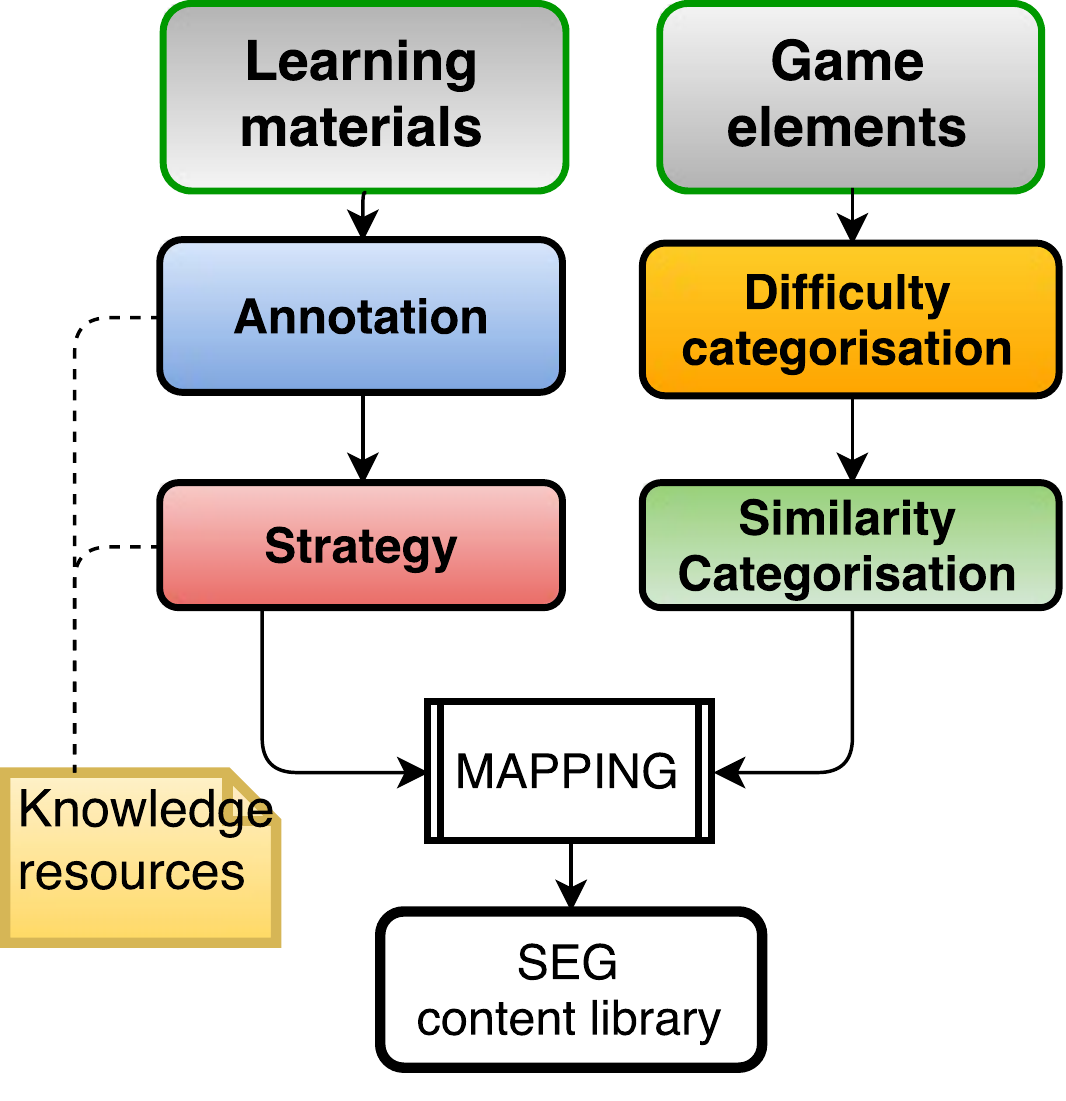}
\caption{An alternative SEG development framework. }
\label{fig:stage1}
\end{figure}

To address the issues mentioned above, we propose an alternative framework for SEG development as illustrated in Fig. \ref{fig:stage1}.
First, learning materials and game elements are in separate spaces. In one hand, annotation takes place to describe education materials naturally from the meta-data retrievable from reliable resources. Then, we need to establish the strategy for presenting them to players, based on their retrieved properties or using the corresponding educational resources. On the other hand, categorisation of game content space consists of a couple of steps. It starts with a difficulty categorisation which groups game content according to the level of challenge. Subsequently, within each of the pre-defined content categories, e.g. difficulty levels, and given a number of education materials, clustering analysis is applied to group similar game content. Hence, the aspects underlying the descriptive learning materials and game elements can guide a developer to use their logic in formulating the mapping between learning materials and game content. The outcome is an SEG content library comprised of playable games for learning.

\subsection{Knowledge Space}

\textbf{Knowledge space} of an SEG refers to all the relevant learning materials consisting of items to be learned by a player. Assuming that no organization had taken place, Belloti~\cite{belloti:task} demonstrates an annotation technique for serious games' tasks which inspires us in structuring learning materials. However, the author employs experts to annotate subjective attributes.
Again, the author fails to assure whether his approach can handle the growing size of learning materials. Especially for serious games, where a large number of education materials are recalled, such as memorizing vocabulary, biology terms, geographical items.

\par
On the other hand, we argue that the ideal properties for learning materials originate from its inherent description provided by reliable education resources (e.g. syllabus) and the representation of the knowledge (e.g. text, image, audio or video). Hence, \textbf{annotation} operates in the natural descriptions of the learning materials with little involvement from experts.
Specifically, the developer must select the relevant properties/attributes based on the recalling purpose and/or their representation within the game. Given the available documented resources, an information retrieval technique --beyond our scope --automates the annotation process. In addition, a simple computer program that measures specifications of a content representation (e.g. text, image, audio or video) also annotates the attributes automatically. Such as, the number of words of the text-based learning material or the length of an educational video.
Altogether, the education content space provides comprehensive detail to initiate the \textbf{strategy} for delivering the learning materials. If no relationship exists (e.g. prerequisites of learning words are recognising the letter and their conjunctions) between education materials, an automated method (e.g. rule-based) establishes the strategy based on the attribute values. Otherwise, a syllabus or a teaching handout can show the strategy explicitly. Consequently, with an established strategy, players will recall the knowledge accordingly.

\subsection{Game content space}

\textbf{Game content space} of an SEG refers to all the playable games generated by an entertainment game engine to facilitate the learning defined via knowledge space for a player. Our approach applies procedural content generation for each element in the game because, ideally, it provides details of the content in the parameters. Given the large space for the generated content, manually identifying the category for the content space is not feasible. Therefore, we adopt a set of steps applied in entertainment game procedural content generation\cite{robert2013learningbased}, including: \textbf{difficulty categorisation} and \textbf{similarity categorisation}. Difficulty categorisation provides games for players with different abilities for playing the game\cite{csiks1990flow}. Meanwhile, similarity categorisation benefits the large space of the content which provides abundant choices of games which support repetitive sessions of learning.

\par
Robert and Chen suggest \cite{robert2013learningbased} that the use of categorized game content can
facilitate eliciting positive affective experience via a proof-of-concept first-person shooter game where content has been categorized via difficulty levels learned from game examples annotated by developers. For specifying difficulty levels, a developer can also adopt a rule-based approach. As a result, the rules are formulated by taking into account a small number of game controlling parameters where developers have to decide the threshold values of those parameters to split content space into proper regions of different difficulty levels. Consequently, content categorisation naturally takes place with the specified difficulty levels. Nevertheless, compared to the aim in \cite{robert2013learningbased}, a different purpose of clustering analysis occurs here. Given the value of $k$ as a total number of chunks of knowledge, the analysis identifies $k$ groups of similar game content for each education material and prevents boredom developing when multiple repetitions of a game session required.

\par
\subsection{Mapping between Knowledge and Game Content Spaces}

\textbf{Mapping} between knowledge and game content spaces is the essential step that deploys each learning material into game content based on their underlying characteristics. An arbitrary or sampling-based mapping may be the simplest method. However, it may promote ineffective learning for different players. In fact, learning and playing in an educational game involves various factors~\cite{dpavlas2010flowSEG}. For instance, arbitrary mapping has a greater possibility to assign an uncomplicated recall materials with difficult games. Hence, a novice player may struggle playing difficult games trying to overcome such challenges. This situation could hinder a player's aim to recall the learning material. In other words, using arbitrary or sample-based mapping can produce imbalanced outcomes for the players.

Therefore, mapping should follow specific rules that produce acceptable deployment of learning materials and produce relatively fair experiences for various types of player.
For now, our strategy employs the developer's intelligence to exploit the in-depth characteristics of each content space. According to the content structures, the crucial mapping rule embeds an education material into a unique cluster of game content from each difficulty level.
Therefore, it can prevent boredom growing when they need to repeat learning the same knowledge.
Additionally, we recommend creating a rather detailed rule set through the following steps.
Let the education materials be a series of learning tasks.
One can identify the situations that elicit different outcomes when learning adjacent, significantly different (e.g. first and last chunk) or correlated knowledge chunks.
Identifying those situations is somewhat abstract; however, it still in the developer's mind.
Essentially, the developer must estimate the specifications of a game cluster that supports an identified condition.
Hence, additional rules can drive a more acceptable mapping concerning the player's experiences.
Using the rule set, we can deploy learning materials and game elements automatically even when both have large spaces.

\section{Case Study: Chem Dungeon for recalling Chemical Compounds}
Using the method presented in Sect.~\ref{sec:method}, it allows a developer to transform an existing entertainment game into an educational game by embedding learning materials. Conceptually, the method should be applicable for combining various learning subjects and various games. Therefore, the next subsections describe an implementation of our method based on and existing game, Chem Fight, including the solutions tackling the practical challenges.

\subsection{Chem Fight}
One of the authors (MP) developed the Chem Fight open sourced under MIT licensing\footnote{accessible online: \url{http://js13kgames.com/games/chem-fight}, and the source code is available online: \url{https://github.com/mpalmerlee/ChemFight}.}, a turn-based game that confronts a single-player versus a Non-Player Character in a chemical compound battle. Whereas, attributes of known 20 atoms from the Periodic Table (PT) and the atom bonding rules construct the gameplay.

\par
Both players have a number of lives (red heart icon), energy (blue flash icon) and Atom Bucks (yellow dollar sign). The following paragraph explains the game mechanics with clarifications\footnote{available online: \url{https://github.com/mpalmerlee/ChemFight}}.

\par
The game consists of a number of rounds until one of the players loses all their lives. Each round consists of a purchasing mode; one turn for the player to defend and another for attacking the NPC. The purchasing mode allows each player to buy atoms from the periodic table at a price specified by its atomic number (e.g. Helium [He] with atomic number 2 costs two Atom Bucks).
On the first turn, one player attacks with a single atom, the defender (the other player) only sees the valence electron count for the attacking atom, thus, it earns a chance to appoint a number of atoms for defence. If the attacking atom creates a chemical bond with one or more of the defender's chosen atoms (a successful defence), the defending player receives rewards composed of a number of Atom Bucks and Energy Units. Otherwise, if there is no known possible compound between the attacking element and any of the defending elements, the attack is successful and the defending player receives a penalty for those unbounded defending elements. In fact, such a rule should discourage players from just defending with every element they own each time. Meanwhile, regaining the unused defending elements costs a decrease in energy. However, if the player has insufficient energy, their health decreases in proportion to the deficit. Once each turn ends, players earn a number of Atom Bucks to allow them to spend on additional elements.

\subsection{Chem Dungeon}
Inspired by Chem Fight, we apply our proposed approach in Sect.~\ref{sec:method} to develop a new SEG: Chem Dungeon which has entirely different game mechanics compared to the original. Indeed, we use the library of education materials and the basic game rule (pairing atoms to create a compound) as the core of Chem Dungeon. Moreover, an existing rogue-like game \footnote{available from \url{http://www.kiwijs.org/}} is employed to represent the game content. Given both spaces are available, the following subsections describe the process details.

\subsubsection{Learning Materials: Chemical Compounds}

The educational game has a purpose in promoting the memorization of chemical compounds for players. For this case study, there are 100 compounds composed of at least two atoms to learn by the players. The textual representation informs a compound's symbol, name and the atoms. For instance,"\textit{2 Hydrogen and 1 Oxygen construct an H2O (water)}" represents the water compound. One of the atoms (e.g. $H$) comprises of a singleton while the other allows multiple atoms of the same type (e.g. $2O$). Meanwhile, a single atom appears as a game object with a text-based atomic symbol, e.g. \textit{O, Ca, C}l. Otherwise, if numerous atoms of the same type are involved it appears as a concatenation of strings between the total atom and the atomic symbol, e.g. \textit{2O, 2H, 2Cl, 6B}.

According to Fig. \ref{fig:stage1}, there are two general steps to proceed. First, given the periodic table data, attributes appointment operates according to the forming atoms and compound representation. Attributes of the forming atoms (atom-1 and atom-2) include \textit{atom-1-number} (discrete),\textit{ atom-2-number} (discrete), \textit{total-types-of-atom} (discrete) and \textit{total-atom} (discrete). Attributes associated with compound and atom representations include: \textit{total-character-symbol-1} (discrete) and \textit{total-character-symbol-2} (discrete). Subsequently, a computer program retrieves necessary data from the periodic table and measures the total characters for the involving atoms, then, annotates the attributes automatically. For instance $CO_2$ comprised of one Carbon and two Oxygen atoms. The annotated values of this compound are \textit{atom-1-number=6} (C), \textit{atom-2-number=8} (O), \textit{total-types-of-atom} $CO_2$ is 2 \textit{(1 C + 1 O)}, the \textit{total-atom} is 3 \textit{(1 C + 2 O)}, \textit{total-character-symbol-1}  and \textit{total-character-symbol-2} are both 1.

Second, with the fact that no correlations exist between compounds, the strategy of remembering them takes into account the properties. In fact, recalling them should be driven by the complexity of each compound. In other words, the more complex the representation of a compound, the more \textit{difficult} it is to memorise. Accordingly, the strategy in our case associates with structuring education materials in a specific order based on the priority of attributes for sorting. Therefore, based on recall priority, compounds are ordered based on \textit{total-types-of-atom}, \textit{total-atom}, \textit{atom-1-number} and \textit{atom-2-number}, \textit{total-character-symbol-1} and \textit{total-character-symbol-2}, respectively. As a result, the easiest compound to remember is H2 (composed of two Hydrogen atoms) and the hardest to recall is CaB6 (formed from one Calcium atom with six Boron atoms).
Hence, the sorted compounds are then represented by the CompoundID attribute which has numeric values from 1 to 100.

\subsubsection{Game Content Space: Rogue-like Maze}

The game content space was constructed from an existing rogue-like and maze game to confirm that it segregates from the learning materials. Henceforth, the categorization and mapping processes become revealing for our demonstration. As an overview, generating game elements using parameter values applies here which consist of \textit{maze-id} (categorical), \textit{enemy-type} (0: random-move enemy, and 1: smart enemy), \textit{total-enemy} (1-5), \textit{total-bullets} (1-5). By default, the game content space counts 48600 different parameter configurations.

\begin{table}\small\sf
\caption{Difficulty Categorisation Rule Set.}
\centering
\label{table:diffrule}
\begin{tabular}{l c c c c}
\toprule
\textbf{Difficulty} & \multicolumn{1}{l}{\textbf{\begin{tabular}[c]{@{}l@{}}enemy-\\ type\end{tabular}}} & \multicolumn{1}{l}{\textbf{\begin{tabular}[c]{@{}l@{}}total-\\ enemy\end{tabular}}} & \textbf{\begin{tabular}[c]{@{}c@{}}total-\\ bullets\end{tabular}} & \textbf{maze} \\ 
\midrule
\textbf{Easy}                    & 0                                                                                   & \textless 4                                                                          & \multicolumn{2}{c}{\multirow{4}{*}{any}}                                         \\ \cline{1-3}
\multirow{2}{*}{\textbf{Medium}} & 0                                                                                   & \textgreater 3                                                                       & \multicolumn{2}{c}{}                                                             \\ \cline{2-3}
                                 & 1                                                                                   & \textless 3                                                                          & \multicolumn{2}{c}{}                                                             \\ \cline{1-3}
\textbf{Hard}                    & 1                                                                                   & \textless 2                                                                          & \multicolumn{2}{c}{}                                                             \\ 
\bottomrule
\end{tabular}
\end{table}

\par
\begin{figure}
\centering
\includegraphics[scale=0.23]{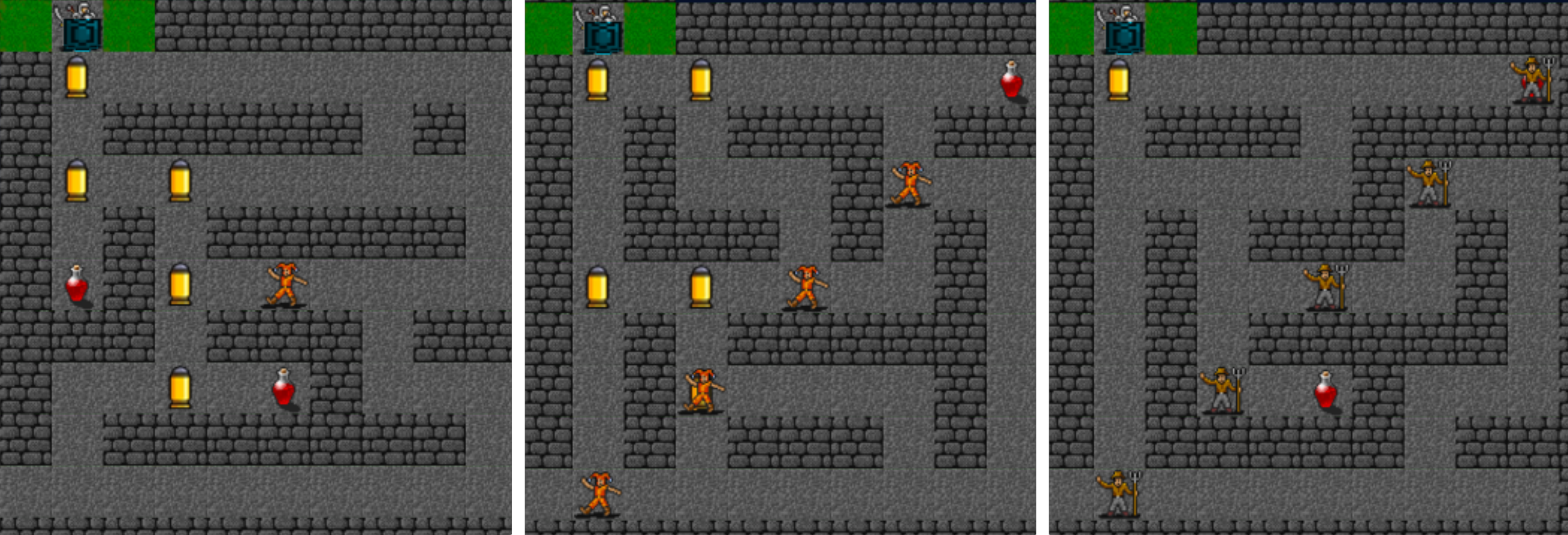}
\caption{Exemplary games of different difficulty levels (left-right): Easy, Medium, Hard.}
\label{fig:difficultLevel}
\end{figure}

In \textbf{difficulty categorisation}, three levels of challenges separate the game content. To our best knowledge, the parameters \textit{enemy-type} and \textit{total-enemy} distinguish the difficulty quite noticeably within the rule set in Table \ref{table:diffrule}.
As a result, 22365 of game content is categorised as \textbf{Easy}, 15660 is \textbf{Medium}-level game content and 10575 of content has a \textbf{Hard} difficulty level. Fig. \ref{fig:difficultLevel} illustrates three different levels of difficulty. The image on the left is identified as an \textbf{Easy} game. Due to this fact, there is merely a single obstacle from one enemy which moves randomly, but the avatar can wander around the maze freely without very much concern being hit by the sole enemy. The image in the middle and on the right are \textbf{Medium} and \textbf{Hard} difficulty levels, respectively. A medium game, provided with four enemies moving randomly allows free movement for the avatar with added challenges to avoid collision with these enemies. Meanwhile, the game content with five Smart enemies demands a high level of tactical practice in decision-making because these enemies are capable of traversing the shortest path to the avatar.

%image of easy medium difficult games

\par
Our goal in the \textit{similarity categorisation} is to provide a selection of similar game content for each learning material. To accommodate this, a clustering analysis builds ($k=100$) clusters of similar game content inside a difficulty group.
Given that \textit{maze-id} parameter do not describe a maze explicitly, five numeric parameters represent it measuring the maze's \textit{total-path}, \textit{total-corners}, \textit{total-intersections}, \textit{total-deadend} and \textit{complexity}.
Aside from this, awareness was raised over a number of challenges: 1) the large size of game content space, and 2) the dynamic size of the content space due to the previously played games. Accordingly, our choice falls to Balanced Iterative Reducing and Clustering using Hierarchies (BIRCH) which is fast and flexible even with very large samples (details available in \cite{zhang1996birch}).

\par
For our case, configuring BIRCH with $k=100$ and setting the branching factor $B=2$ constructs a binary tree of game content space for easier visualisation. Subsequently, the BIRCH operates to search for an optimum threshold value $T$ which identifies 100 clusters with the highest silhouette score as an evaluation measure. The result of BIRCH on game content space under normalised values attests to Low, Medium and High difficulty groups using a threshold $T_l=T_m=T_h=0.02$ to reach the highest silhouette of 0.23, 0.2 and 0.23, respectively.

\par
Overall, 300 clusters identified and equally divided into 3 difficulty levels are ready for deployment with the education materials.

\subsubsection{Mapping: A Rule-based Approach}

Previous steps successfully arranged the chemical compounds from the simplest to the most complex to memorize, and the game clusters carry details including total game content, linear sum of each parameter, the sum of square of each parameter and the centroid points of the cluster. In addition, statistics for each cluster can serve as the game content description.

\begin{figure}
\centering
\includegraphics[scale=0.7]{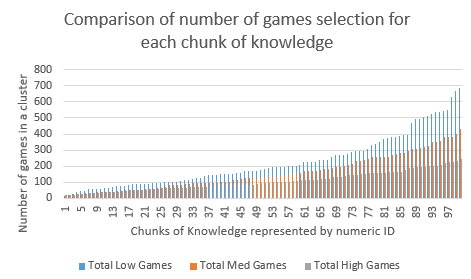}
\caption{Mapping result in terms of number of games in a cluster.}
\label{fig:mapN}
\end{figure}

\par
Our deployment strategy operates a rule-based method. Given the specifications found in learning material and content space, the following crucial rule applies: a compound deploys into three unique content clusters, all from different difficulty levels. Indeed, this rule ascertains no duplicates of game content for multiple compounds. However, additional criteria ensures an appropriate mapping based on our notion of possible learning conditions between \textit{simple} versus \textit{complex} compounds. Essentially, \textit{difficult} or complex compounds may need a higher number of games to play with.
In fact, the likelihood of failures to recall complex compounds may be higher than the easier ones. Thus, a higher frequency of repetitions may transpire for learning complex compounds. As a consequence, a slight difference in the game elements for recurrence of memorization may accustom the player to those games without the fear of boredom growing.
Therefore, the player may have a \textit{wider space} for focusing on the learning goal. Given these expectations, the cluster details resemble those aforementioned conditions including the quantity of game content (represents the number of repetitions) and the sum of standard deviations of game content features (represents the variety of games) under non-normalized parameter values.
The following pseudo-code shows the deployment rules in practice.

\begin{enumerate}
\item Assign education materials with string ID $E_j$, where $j: 0,1,...n$ (n is the learning materials size).
\item Within each cluster (\textit{j: 0,1,...n}) of each difficulty level (\textit{i: 0,1,...m} total difficulty levels): count games ($N^i_j$), sum the standard deviations of parameters ($S^i_j$) and assign the game content with string ID $G^i_j$.
\item Sort clusters within each difficulty level based on the value of \textit{N} (ascending) and \textit{S} (descending), respectively.
\item Create pairs of $[E_j, G^i_j]$, where j:0-n and i:0-m, enabling an education material gets a cluster of game content from each difficulty level.
\end{enumerate}
%\begin{algorithmic}[1]
%	\STATE Get CompoundID ($E$) from the sorted compounds
%	\FOR{$i=0$ to $i=totalDifficulty$}
%	\FOR{$j=0$ to $j=totalClusters$}
%		\STATE COUNT game elements, stored in $n^i_j$
%		\STATE Compute Sum of Standard Deviations of parameters of game elements, stored in $s^i_j$
%		\STATE ASSIGN gameID, stored in $G^i_j$
%	\ENDFOR
%	\ENDFOR
%	\STATE SORT	clusters within each Difficulty Level based on $n$ then $s$
%	\FOR{$i=0$ to $i=totalDifficulty$}
%	\FOR{$j=0$ to $j=totalClusters$}
%		\STATE CREATE a new mapping array $[E_j, G^i_j]$
%	\ENDFOR
%	\ENDFOR
%\end{algorithmic}
\par
\begin{figure}
\centering
\includegraphics[scale=0.7]{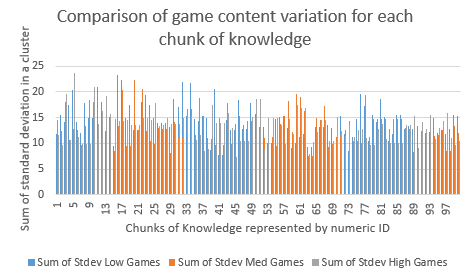}
\caption{Mapping result in terms of the sum of standard deviation of game content parameters in a cluster.}
\label{fig:mapS}
\end{figure}

Mapping priority starts with the number of games in a cluster (Fig. \ref{fig:mapN} depicts result details regarding the number of games for each compound) and is followed by the standard deviation of the cluster (Fig. \ref{fig:mapS} shows deployment details with respect to variations of the game content). SQL-based tables store the mapping result and both content space details.

%graphs of mapping
\subsection{SEG Game Engine}
\begin{figure}
\centering
\includegraphics[scale=0.48]{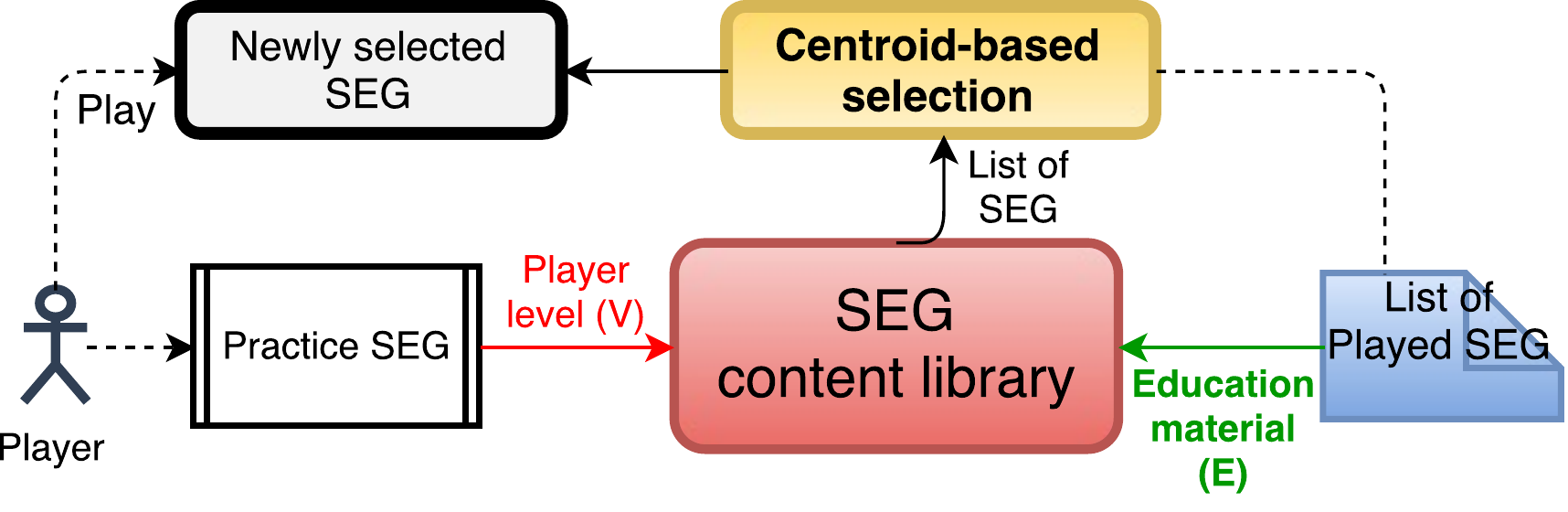}
\caption{Procedure in an SEG game session.}
\label{fig:stage1-gen}
\end{figure}

Given the structure of the content of a relatively new game, a game engine should properly situate different players accordingly. Fig. \ref{fig:stage1-gen} shows several stages in the SEG game engine. Initially, a new player should accustom him/herself with the game-play in the \textit{practice game session} which contains the educational game with dummy learning materials. Meanwhile, an existing player may enter the practice game session for updating his/her Player Level. As a matter of fact, this session estimates the mastery level (denoted as $V$) of the player with the game based on his/her \textit{score achievement}. Whereas, the mastery level $V$ corresponds to the difficulty level of the game content.

In principle, the score originates from the player's game actions which consist of positive ($a^+$) and negative ($a^-$) actions. Logically, positive game actions increase $score$ such as through successful navigation or accurate shots while negative game actions reduce $score$, for instance, a failed navigation or failed battle. In addition, various weights (if known by the developer) on particular actions may yield a more accurate scoring. Equation $score = \sum_{i}^{k}\alpha_i a^+_i - \sum_{i}^{l}\beta_i a^-_i$ provides the basic formula for scoring, where $a^+_i$ be the $i^{th}$ positive game action and $a^-_i$ be the $i^{th}$ negative game action. A value of $k$ counts the number of positive game actions while $l$ measures the total negative game actions. Values of $\alpha_i$ and $\beta_i$ set the $i^{th}$ weights for positive and negative game actions, respectively. Then, the threshold values of $score$ categorise a player into a particular level $V$.

Initially, a new player starts playing the game with the first education material ($E$). Meanwhile, an existing player may progress the educational game according to his/her game session record (List of Played SEG). Based on $V$ and $E$, the generation engine searches through the \textit{content library} for a specific education material, and the corresponding game content cluster, as game content candidates. For a new player, the candidates are \textit{all} games in the selected cluster. On the other hand, the played game content is excluded from being a candidate. Subsequently, a \textit{centroid-based selection} chooses the closest game to the centroid $x_m$, measured by \eqref{eq:centroid}, of the pool as the newly selected game. Whereas, $x_i$ be the $i_{th}$ game content in the pool and $n$ be the number of game content candidates.
\begin{equation} \label{eq:centroid}
x_m = \frac{\sum_{i}^{n}x_i}{n}
\end{equation}
Finally, the game engine generates the \textit{newly selected game} composed of parameters incorporating the \textit{education-id} (based on the value of $E$) and the value of $V$ which associates the game content features.

\subsection{Game Mechanics}
\begin{figure}
\centering
\includegraphics[scale=0.40]{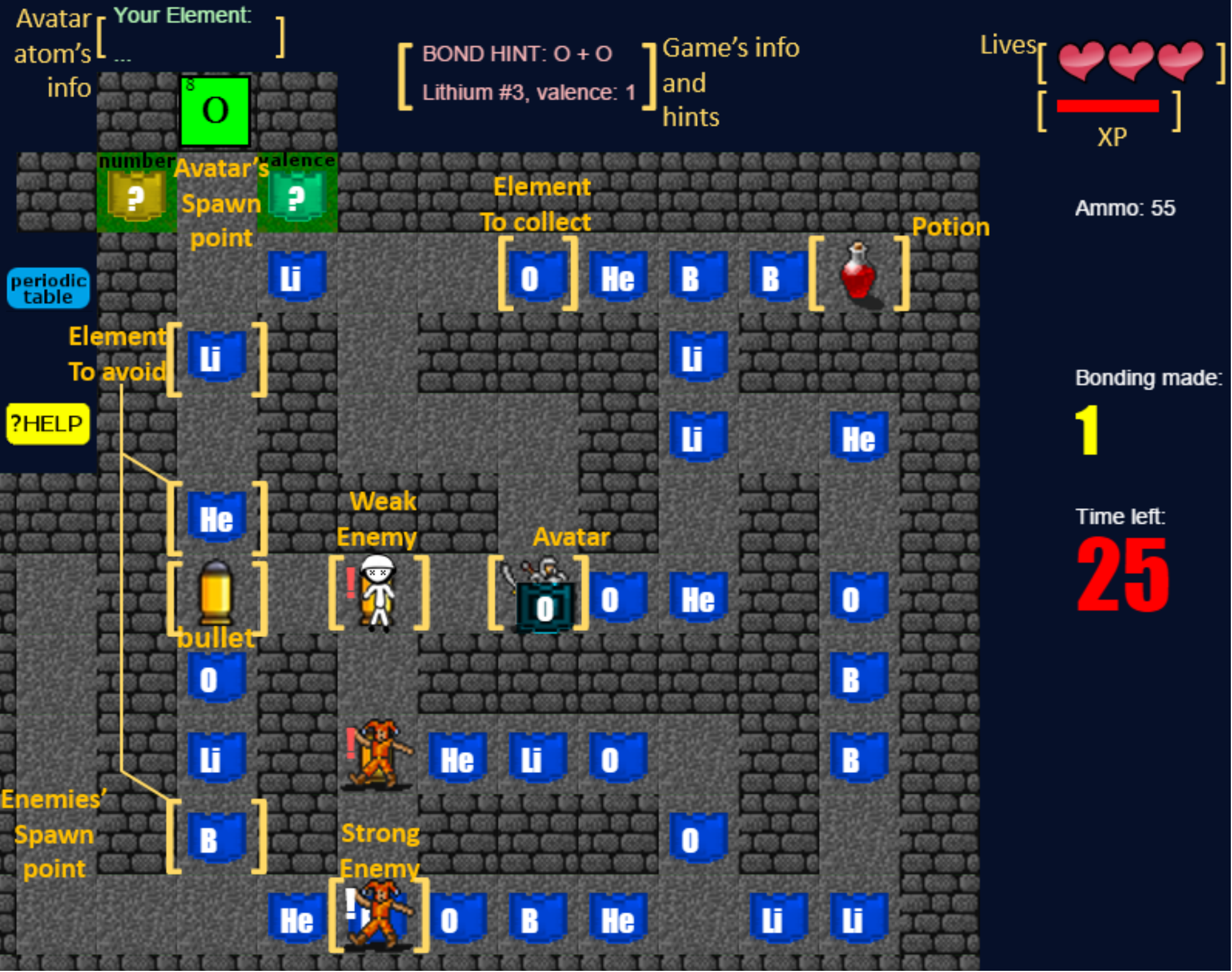}
\caption{The Chem Dungeon layout.}
\label{fig:cf}
\end{figure}

This section demonstrates the game-play of the newly-developed SEG as observed in Fig. \ref{fig:cf}. The game field consists of pathways and walls that form a maze with intersections and cul-de-sac. An exit gate, initially closed, is hidden at the bottom-right of the maze. Actors in the game consist of an avatar and a number of opponents, each with a spawn point. The avatar carries an atom within its shield of which information is shown near its spawn point (top-left corner), a button to open the periodic table and a \textit{Help} button to pause the game and show mission objectives. Meanwhile, information regarding a compound-forming result or an atom properties are at the top-centre of the game arena. The right side of the game (from top to bottom) contains: lives (heart icon), experience in a red bar, ammunition (number), bonds made (number) and the remaining time (90 to 0 seconds).
Inside the maze, bullets (yellow object), atom objects (blue object) and live potion (red object) are collectible for the avatar. Each bullet collected adds some ammunition for the avatar. A live potion can restore the avatar's life to full.

The objectives of the game comprise of the forming of a compound and entering the exit gate within the 90-second time limit. Initially, the avatar starts from its spawn point while the enemies start in the diagonal paths of the maze (bottom-left to top-right). The avatar can walk in 4-degrees of freedom: left, down, right, up controlled by keyboard keys \textit{a}, \textit{s}, \textit{d}, \textit{w}, respectively. When exploring the maze, the avatar should avoid colliding with an enemy or an atom object.  In fact, it will surrender one life when colliding with an incorrect atom object or a normal enemy. Luckily, shooting an atom object opens a path due to the shot atom changing its position to another empty tile. Meanwhile, shooting an enemy transforms it to a weak mode (white-coloured character). A weak enemy re-spawns back to its home when crashing with the avatar, thus, opening another clear route. Accordingly, the avatar can seek and assemble the correct atom object which creates a compound. At this point, an educative message pops up which contains information concerning the chemical compound. Indeed, this game state should encourage players to read and retain knowledge in their memory. When the avatar has collected the correct atom object ten times, the exit gate reveals to open. Finally, by entering the exit gate, the avatar gets a \textit{Victory}. Otherwise, losing all lives or out of time issues a \textit{Defeat}.

The following illustrates some helpful hints for players to play the game. Although each game contains different atom objects, its aim is to form one compound (repeatedly). Novice players can adopt a trial-and-error strategy and are fully aware not to lose all their lives. Therefore, the player ought to actively read the text message at the top-centre position of the game which holds the latest result for the compound-forming attempt. Meanwhile, if only one life remains, a player can regain full lives by collecting a potion. Or, similarly, by accumulating experience (XP) bars through accurate shots and hit weak enemies. Once the XP reaches a full bar, one additional life replaces it. However, such an endeavour should consider the remaining bullets/ammunition and the 90-second time limit. These restrictions impede players abusing such tactical practices merely for entertainment while disregarding the goal of playing the game: memorising compound forming.

\section{User Test}
Developing an educational game using the method presented in this paper can produce a 'new' game, due to the mix of education materials and game content. Therefore, a survey containing the SEG allows players to play the game and report their experiences. The survey opens only for players at least 18 years old and computer literate.

\begin{figure}
\centering
\includegraphics[scale=0.45]{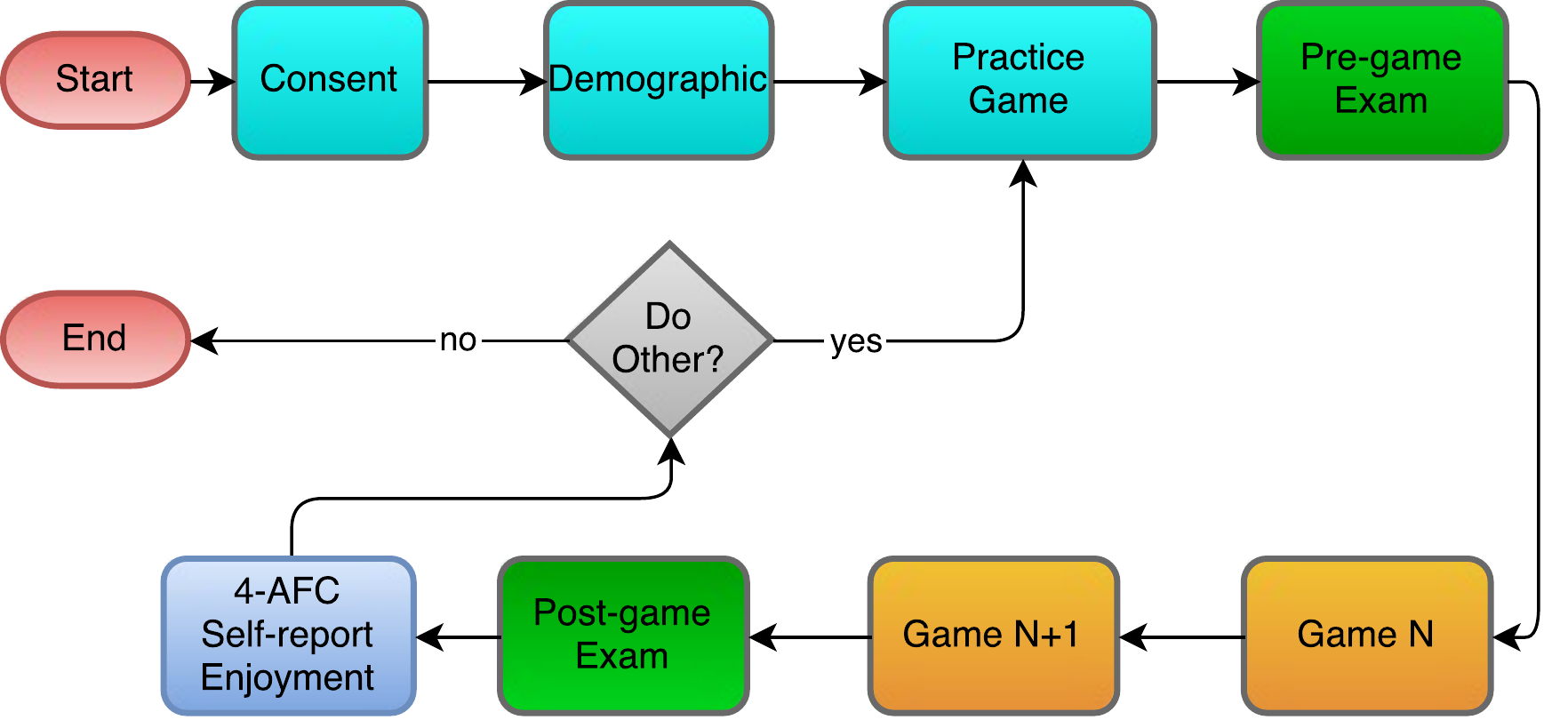}
\caption{Procedure for user survey.}
\label{fig:survey}
\end{figure}

Fig. \ref{fig:survey} depicts the procedures for the survey commencing with a Consent form, Demographic form, Practice (training) session and Pre-game Exam (randomly chosen learning materials). Afterwards, players play a pair of games, each with a single education material contained in the Pre-game Exam and a difficulty level for the game content according to his/her level measured from the Practice game session. Following each pair of games a questionnaire is provided for players to report \textit{fun}/enjoyment from the latest played games and a Post-game Exam. In addition, each game session produces a log of gaming activities for further analysis.
The consent form confirms a player's participation in the survey. Meanwhile, the Demographic form records participant data, including age, location, player-id, email address and a unique code for players to re-enter the survey. A 4-afc questionnaire expects a player to compare his/her enjoyment between both games~\cite{pedersen2010modeling, 4afcjakel2006spatial}. Question wording for the reported enjoyment appears as follows: a) \textit{Game N+1 is more FUN than Game N}, b) \textit{Game N is more FUN than Game N+1}, c) \textit{Both Games are FUN} and d) \textit{NONE of the Games are FUN}.
Meanwhile, Pre and Post-game Exams employ Multiple Choice Question (MCQ) design~\cite{papastergiou2009digital, manske2005c}.

Subsequently, a player may revisit the training session if he/she requires improving his/her gaming ability before continuing to the next section of the survey. Alternatively, he/she may opt to directly play a new pair of games initialised by completing another pre-game exam, or just quit the survey.

\subsection{Data Analysis}
The survey ran for three months and the 50 players participating were allowed to play several pairs of games, hence, 540 reports obtained. Ten games were played on average. Four players played only a pair of games while 85\% played between 4 to 14 games. One player played and reported 15 pairs of games. Regarding the reported experiences, a z-test will infer the difference between contrasting affective experiences, e.g. proportion of enjoyment versus no enjoyment.

In terms of reporting Enjoyment, 352 reports confirmed that the games were entertaining while 188 games reported they were not enjoyable. Table~\ref{table:ztest1} summarises three z-tests evaluating $H_0$ against $H_a$. The null hypothesis $H_0: \pi = 0.5$, where $\pi$ indicates the proportion of FUN reports.  Given the 0.01 significance level, two z-tests reject the null hypothesis while 99\% confident the proportion of FUN reports (0.652) is greater 0.5 proportion.

By looking into play-logs, we observe that there are slight differences between various gaming activities that separate the reported Fun and Not Fun. This is due to the fact that the Fun experience is a very subjective matter. One player may feel 'entertained' due to the game content fits his skill, while another player may feel an enjoyment when the game content choice is more difficult to play for his/her skill. This is a factor among many others as it is common that different players could have various perceptual/cognitive experience in response to the same stimuli. Moreover, the affective experience may change overtime or known as \textit{concept drift}~\cite{robert2013learningbased}. To be more precise, the questionnaire should have contained a thorough questions that accommodate various aspects of enjoyment~\cite{fu2009egameflow}. The observation also suggests that the game content adaptation should be introduced in SEG development. 

\begin{table}\small\sf
\caption{Z-test on Proportion of Gained Enjoyment.}
\centering
\label{table:ztest1}
\begin{tabular}{l l l l}
\toprule
\multicolumn{4}{c}{\textbf{Z-test}, $H_0$ against:} \\ 
\midrule
\textbf{Indicators} & \textbf{$H_a: \pi \neq 0.5$} & \textbf{$H_a: \pi > 0.5$} & \textbf{$H_a: \pi < 0.5$} \\ \hline
p-value& 0.00000 & 0.00000 & 1 \\ \hline
99\% conf.& 0.59-0.74 & 0.6-1.0 & N/A \\
intervals& & & \\ \hline
$H_0$ status& Rejected & Rejected & Rejection\\
 & &  & Failed \\ 
\bottomrule
\end{tabular}
\end{table}

Regarding the learning performance of the players, each question item in an exam represented a learning material. Thus, pre and post-game exams produced binary values indicating prior knowledge and recalling results, respectively. The difference in scores between pre and post-game exams produced three types of learning performances, i.e. unchanged, improvement and decay. However, only the unchanged and improved learning performances will be used due to the negative score (decay) likely to be produced from arbitrary answers or random guess~\cite{bao1999dynamics}. In addition, game sessions exist containing known learning material that the exam design was not designed to measure for the improvement of learning in such cases. Therefore, 309 reports involving not known prior knowledge were divided into 219 game sessions which helped players improve their learning performances, while 90 sessions failed to improve learning performance. For this case, the same z-tests operate using the same values for the null hypothesis and alternative hypotheses while $\pi$ indicates the proportion of improved Learning. Table~\ref{table:ztest2} summarises three z-tests results. Given the 0.01 significance level, two z-tests reject the null hypothesis while 99\% confident the proportion of Improved Learning reports (0.694) is greater 0.5 proportion.

Furthermore, we investigate the recorded gaming activities corresponding to \textit{learning} and  \textit{not learning} outcome and find gaming activities seem correlated to learning outcome. In general, a game session where players recalled most of the education materials has more gaming activities than a game session where players only recalled few or no the education materials. In fact, the total time spent in reading the successfully collected compound corresponding to \textit{learning} actions take around 15 seconds on average. In contrast, the \textit{not learning} actions always take less than three seconds. Overall, the total actions in \textit{learning} game sessions over the \textit{not learning} in-game activities have been doubled approximately. This is due to the fact that the goal of this educational game is designed to collect as many  bond-able atoms as possible, which properly reflects learning.

\begin{table}\small\sf
\caption{Z-test on Proportion of Improved Learning.}
\centering
\label{table:ztest2}
\begin{tabular}{l l l l}
\toprule
\multicolumn{4}{c}{\textbf{Z-test}, $H_0$ against:} \\ 
\midrule
\textbf{Indicators}& \textbf{$H_a: \pi \neq 0.5$} & \textbf{$H_a: \pi > 0.5$} & \textbf{$H_a: \pi < 0.5$} \\ \hline
p-value& 0.00000 & 0.00000 & 1 \\ \hline
99\% conf.& 0.65 to 0.73 & 0.66 to 1.0 & N/A \\
intervals& & & \\ \hline
$H_0$ status& Rejected & Rejected & Rejection\\
 & &  & Failed \\ 
\bottomrule
\end{tabular}
\end{table}

The statistical evidence shown in Table \ref{table:ztest2} confirms that the Chem Dungeon game is considered successful from the players' perspective regarding their learning and enjoyment, which is consistent with Pavlas' testimony~\cite{dpavlas2010flowSEG}.

On the other hand, we also look into the relationship between learning outcome and affective experience reported by the survey participants. Table~\ref{table:multixp} summarizes such information collected from all the game sessions. It is evident from Table~\ref{table:multixp} that our SEG allows more players to gain positive learning outcome and Fun together as there are 154 out of 309 falling into this category. This clearly demonstrates that the use of separate content spaces and a proper mapping proposed in our framework may lead to an SEG that fits all the characteristics described by Abt in 1970s~\cite{abt:sg}: in a serious game, learning may be primary  but other experiences involved should not be overlooked! Furthermore, serious games involve learning and entertainment dimension as a unity during game sessions~\cite{SG:michaud2008serious, SG:jantke2010toward, zyda, ritterfeld2009serious}. Recent research by Pavlas found that enjoyment arising from the playing activities may affect the learning of a player in a serious game~\cite{dpavlas2010flowSEG}. While the learning in serious games is a primary objective that any players have to achieve, our work emphasizes the importance of enjoyment (entertainment). Overall, the experimental results reported above indicate that, to a great extent, our game content and rules may elicit positive affective experience and many players gain such enjoyment when they engage in learning via game playing.

\begin{table}\small\sf
\caption{Survey Results: Learning Outcome vs. Affective Experience.}
\centering
\label{table:multixp}
\begin{tabular}{lll}
\toprule
       & NotLearning & Learning \\ 
\midrule
NotFun & 42          & 65       \\ \hline
Fun    & 48          & 154      \\ 
\bottomrule
\end{tabular}
\end{table}

\section{Conclusion}
We have successfully developed Chem Dungeon using a strategy that combines education materials and an entertainment game. Retrieving inherent details of the learning materials demonstrates advantages in two regards. First, the learning materials has a natural description held by the attributes enabling a developer to organize them semantically. In the second, computer programs can automatically annotate those attributes with little interference from experts and with a concern merely for the learning materials' size. On the other hand, the procedurally generated game elements in our approach unlock another route towards rapid development of SEG in which categorization becomes automated using a combination of rule-based approach and machine learning. Hence, those detailed descriptions underlying both content spaces facilitate a developer in establishing the mapping rules based on his/her knowledge. Moreover, the new educational game we have developed using our method has shown reasonable results in supporting players' learning and entertaining them.

It is worth stating that our current case study based on Chem Fight is subject to limitation.
Chem Fight was designed purely for SEG and therefore many of its game mechanics were designed by taking the properties of atoms and compounds into account. Nevertheless, the Chem Dungeon emphasizes a PCG-based SEG which tolerates the recalling type of learning. Hence, it opens opportunities to adjust the method for a different type of learning specified by the knowledge. In addition, the two-space structure could be a baseline for research in procedural serious game-play generation wherein attributions are concerning learning materials, game elements and game rules.

In our ongoing research, we are further developing the game to enable it to predict players' experiences via gaming data, which would lead to a corresponding adaptation method for personalized learning in the SEG. In addition, we are also going to investigate our proposed approach to new SEG development by combining PCG-based entertainment game platforms and appropriate learning materials.

\bibliographystyle{icstnum}
%\bibliography{cref}
\end{document}